%% file: main.tex
\pdfoutput=1

\documentclass[11pt]{article}

\usepackage{EMNLP2022}

\usepackage{times}
\usepackage{latexsym}

\usepackage[T1]{fontenc}

\usepackage[utf8]{inputenc}

\usepackage{microtype}

\usepackage{inconsolata}

\usepackage{dsfont}
\usepackage{color}
\usepackage{amssymb}
\usepackage{MnSymbol} 
\usepackage{graphicx}
\usepackage{amsmath}
\usepackage{amsthm}

\title{Measuring the Measuring Tools: An Automatic Evaluation of Semantic Metrics for Text Corpora}

\author{George Kour\Thanks{denotes equal contribution.} , Samuel Ackerman\footnotemark[1]\\ {\bf Orna Raz, Eitan Farchi,  Boaz Carmeli,  Ateret Anaby-Tavor}\\
         IBM Research AI \\ \texttt{\{gkour, samuel.ackerman\}@ibm.com}\\
         \texttt{\{ornar, farchi, boazc, atereta\}@il.ibm.com}}

\begin{document}
\maketitle

\begin{abstract}

The ability to compare the semantic similarity between text corpora is important in a variety of natural language processing applications.
However, standard methods for evaluating these metrics have yet to be established.
We propose a set of automatic and interpretable measures for assessing the characteristics of corpus-level semantic similarity metrics, allowing sensible comparison of their behavior.
We demonstrate the effectiveness of our evaluation measures in capturing fundamental characteristics by evaluating them on a collection of classical and state-of-the-art metrics.
Our measures revealed that recently-developed metrics are becoming better in identifying semantic distributional mismatch while classical metrics are more sensitive to perturbations in the surface text levels.
\end{abstract}


\section{Introduction}

While there has been a long-standing interest in developing semantic similarity metrics\footnote{In the context of this paper, a metric is a measure of difference (gap) in the general sense, and may not necessarily satisfy the properties of a metric in mathematical terms.} \cite{rayson2000comparing}, measuring how close two text corpora are remains an open problem \cite{pillutla2021mauve}.
Specifically, the recent advances in generative language models have led to an increased interest in the study of content similarity between human and generated language, as a mean for comparing the quality of generative models \cite{mille2021automatic, gehrmann2022gemv2}.

While one can reasonably measure the semantic distance between two individual sentences (e.g., by calculating the cosine distance between the sentence embeddings), measuring the dissimilarity between two text corpora remains a challenge \cite{naeem2020reliable}.
Corpus-level metrics seek to assess semantic similarity at the group level, for instance, assessing generated text fidelity, diversity, and coverage compared to the reference corpus \cite{sajjadi2018assessing}.
Thus, one common approach for measuring the semantic dissimilarity between two corpora is to compare the densities of their sentences in the embedding space
\cite{pillutla2021mauve}.

However, there are no standard automatic procedures for evaluating the precision and robustness of such similarity metrics.
The semi-manual standard approach is to correlate the results of these metrics for human judgement.
However, leveraging manual human judgements to construct numeric metrics has significant weaknesses.
As we explain in Section~\ref{sec:literature}, human judgements are expensive to obtain, are difficult to aggregate consistently from individual text instances into a corpus-level metric in a way that reflects all relevant aspects of the texts, and can be subjective and non-robust.
 


Therefore, in this paper, we adopt a middle ground between validating the metric against human judgement on real data and evaluating the metric with synthetic distributions by building "controllable-distance real data corpora" (Section~\ref{sec:ksc}).
By precisely controlling the content of test corpora, we devised a unified evaluation of desired metric characteristics on real data.
This technique allows aggregation of many small-difference judgements that should correspond to what a human would logically decide, to evaluate the distance metric overall in terms of desirable properties.
The middle ground thus attempts to reflect human logical judgement in an inexpensive way, while avoiding some of the weaknesses described, such as lack of consistency.


%


To summarize, our contributions are as follows.
First, we present a text similarity evaluation measures that allows researchers to compare the robustness of their newly proposed metrics against existing metrics using the same standards.
Second, we evaluate classical and state-of-the-art similarity metrics and show that our benchmark performs well in capturing their known properties.
Finally, we provide a pip-installable Python package to compute an extensive set of text dissimilarity metrics, using a unified and simple API\footnote{https://github.com/IBM/comparing-corpora}.



\section{Literature Review}
\label{sec:literature}

The most widely-used method to compute the quality of text similarity metrics investigates the correlation between the scores given by the metric and human judgements. 
However, human judgement, even on the sentence level, has several shortcomings, mainly that it is expensive and can be inconsistent and subjective \cite{popescu2003experiment, lin2004orange, graham2017can}.
Also, superficial aspects of the sentences, such as text length or syllables per sentence, may influence human judgements of the semantic similarity \cite{novikova2017we}.
Furthermore, though humans may be able judge the relative similarity of a pair of \textit{sentences}, they are usually limited in their ability to make large-scale assessments of a similar type when comparing two \textit{corpora} (i.e., two distributions of sentences) consistently and reliably.

In an attempt to standardize metric evaluation, several competitions and standard datasets containing compared data and human assessment were created for specific tasks, such as translation \cite{guo2018meteor++, mathur2020results}. 
However, there is currently a lack of benchmarks against which to assess the semantic similarity between corpora.



Text similarity metrics can be thought of as belonging to several broad and overlapping classes (see e.g., \citealt{wang2020measurement}), which partially depend on the form of the text representation (e.g., token-based or vector embedding).
Here, we investigate metrics from three of these classes, comparing corpora based on these aspects: \emph{lexicographical} (statistical properties of words and tokens), \emph{ distribution} ( densities of sentences represented in the embedding space), and \emph{discriminatability} (ability to classify sentences as belonging to one corpus or the other).
The metrics we use are summarized in Table \ref{tab:similarity_metrics}.

\begin{table*}
\centering
\begin{tabular}{lll}
\hline
\textbf{Type} & \textbf{Metric} & \textbf{Measures}\\
\hline
Lexicographical  & CHI ($\chi^2$) \cite{kilgarriff2001comparing} & Word/Token count comparison. \\
Statistics   & ZIPF \cite{holtzman2019curious}    & Unigram rank-frequency statistics. \\
\hline
            & FID \cite{heusel2017gans} & Wasserstein distance between densities. \\
Distributional & PR \cite{sajjadi2018assessing}& Assessing distributional precision \& recall. \\
        & DC \cite{naeem2020reliable} & Estimating manifolds density and coverage. \\
        & MAUVE \cite{pillutla2021mauve} & Quality \& diversity via divergence frontiers.\\ \hline 
 Discriminative & CLASSIFIER \citeyearpar{lopez2016revisiting} & Classifiability   between reference and target.\\
        & IRPR \cite{zhao2017learning}  & Average distance between closest samples pairs. \\
\hline
\end{tabular}
\caption{
Summary of investigated text similarity metrics.}
\label{tab:similarity_metrics}
\vspace{-1.5em}
\end{table*}

\paragraph{Lexicographical Statistics}
These methods have been developed to compare various distributional properties of target text $Q$ with respect to the reference samples $P$, based on some statistic measures $T(P)$ and $T(Q)$, operating on the surface text level, e.g., sentence, words, word-parts, tokens, etc. 
Such commonly-used measures include resemblance in vocabulary distribution \cite{kilgarriff2001comparing}, likelihood of repetition \cite{pillutla2021mauve}, and $n$-gram matching \cite{papineni2002bleu}.
However, these metrics tend to be overly sensitive or easily misled by adversarial samples or text peculiarities. 
In general, $\chi^2$-based metrics calculate distance between observed and expected frequencies of categorical variables.
The metric in \cite{kilgarriff2001comparing}, denoted here as \textbf{CHI}, calculates $E$, the average (between $P$ and $Q$) frequencies of the $n$ most common tokens in the combined vocabulary of $P$ and $Q$, then sums the $\chi^2$ statistics comparing each of $P$ and $Q$ to the expected $E$, across tokens.  Here, for both CHI and ZIPF, below, we use the top $n=5000$ tokens.

\noindent
In contrast, the \textbf{ZIPF} metric \cite{holtzman2019curious} compares the use of vocabulary using Zipf’s law, which suggests that the frequency of a given word in human text is inversely-proportional to its frequency rank.
The Zipfian coefficient is fitted on a given corpus and the further it is from $1$, the more the observed corpus differs from the `ideal' theoretical distribution \cite{holtzman2019curious}. 
We can thus use $|z_P-z_Q|$ as a distance metric between corpora $P$ and $Q$.

\paragraph{Distributional Metrics}
These metrics are based on quantifying the distributional relationship between the reference and target corpora in the embedded vector space, thereby capturing semantics beyond superficial token-level statistics.
Here $P$ and $Q$ denote the reference and target corpora in the embedding space. 
Given samples from these, we can use the sample density estimates $\hat{P}$ and $\hat{Q}$ to approximate the true unknown corpus population distributions $P$ and $Q$.

\noindent
The Fr\'{e}chet Inception Distance (\textbf{FID}, \citealt{heusel2017gans}) is computed by fitting a continuous multivariate Gaussian to the $P$ and $Q$, and then calculating the Wasserstein-2 distance between them. 
However, FID is sensitive to both the addition of spurious modes as well as to mode dropping \cite{lucic2018gans}.
Also, while FID is able to detect distributional distances in the high-dimensional space, it cannot shed light upon the nature of this distance.

\noindent
Due to these weaknesses of FID, we additionally consider a metric denoted \textbf{PR} (precision and recall) proposed in computer vision \cite{sajjadi2018assessing, kynkaanniemi2019improved}, which is inspired by the notion of precision and recall.
Intuitively, the precision captures the average resemblance of the individual target samples to the reference set, while the recall measures how well the target samples "cover" the full variability of the reference samples.
To obtain a single distance value using the method in \cite{kynkaanniemi2019improved}, we calculate the $F1$ measure based on the returned precision and recall, denoted here by PR.

\noindent
\citet{naeem2020reliable} proposed an improved estimation of these precision and recall notions by mitigating the overestimation of manifolds caused by outliers and underestimating the similarity when the target and reference are taken from the same distribution.
Similarly to PR, we calculate the $F1$ to obtain a similarity value using this method, denoted as \textbf{DC}\footnote{To calculate both PR and DC, we employed the implementation provided in the \emph{prdc} Python package.}.

\noindent
\textbf{MAUVE} \cite{pillutla2021mauve} is a recently-developed metric that estimates the gap between human and generated text by computing the area under the information divergence frontiers in a quantized embedding space using the KL-divergence\footnote{We used the \emph{mauve-text} Python package for calculating MAUVE as well as ZIPF.}.

\noindent
\paragraph{Discriminatability Metrics}
Similar to the distributional metrics, discriminative metrics calculate the distance using the embedding of the individual sentences in the two corpora.
However, they do not aim to specifically capture the overlap between the distribution induced by the compared corpora. Rather, they calculate the relationship in classification terms, i.e., to what extent can sentences in one corpus be distinguished from the sentences in the other corpus, using a discriminative model.\\
\textbf{CLASSIFIER}: Following \citep{lopez2016revisiting}, we measure the similarity between corpora using a binary classifier. We used SVM \citep{cortes1995support} trained on samples of both source corpora to predict corpus membership in a test set of unseen samples.
A higher test accuracy indicates higher inter-corpora distance.

\noindent
While CLASSIFIER is a model-based metric that uses the entire corpus distribution, \textbf{IRPR} (information-retrieval precision and recall) is an example of an instance- (individual sentence) based corpus distance metric.
Inspired by \citet{zhao2017learning}, we calculate the dissimilarity between corpora as follows.
For each embedded sentence in corpus $A$, we find its closest neighbor in $B$ by cosine distance.
The average of these distances is then computed to find the "precision" value.
The same procedure in reverse, from $B$ to $A$, gives the "recall" value.
We calculate the $F1$ score of the recall and precision to obtain a single value.
Note that the CLASSIFIER metric is used to represent model-based discriminative approaches, while IRPR is used to represent instance-based discriminative methods.

\noindent
The values calculated by CHI, IRPR, PR, DC and Mauve capture the similarity rather than the distance between two corpora (for all metrics $v\in[0,1]$). To make these metrics represent distances, we take $1-v$.

\noindent
Our model selection was based on considering the trade-off between embedding quality and calculation time. 
The code as well as the scripts to reproduce the experiments are available online.\footnote{https://github.com/IBM/meme}


\section{Known Similarity Corpora \label{sec:ksc}}

\begin{figure}
    \centering
    \includegraphics[width=1\columnwidth]{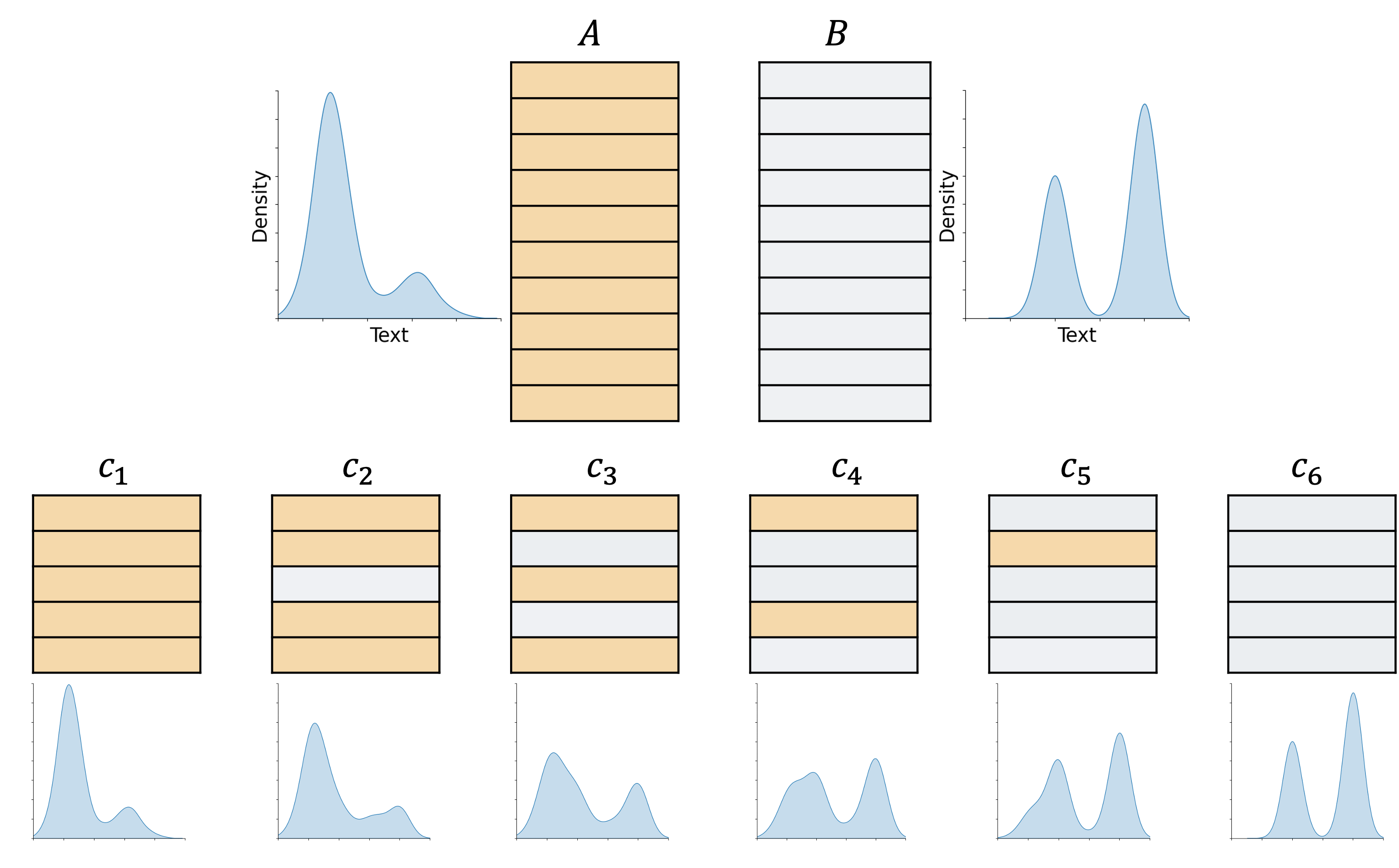}
    \caption{Construction of a $k=6$ known similarity corpora (KSC) collection from source corpora $A$ and $B$. 
    The corpus $c_i$ is constructed by drawing $n\left(\frac{k-i}{k-1}\right)$ and $n\left(\frac{i-1}{k-1}\right)$ samples from $A$ and $B$, respectively. The adjacent densities denote the distributions of the source and KSC set corpora.}
    \label{fig:ksc}
\end{figure}

Most of the metric quality measures we propose are primarily based on the notion of \emph{known-similarity corpora} (KSC) introduced by \citet{kilgarriff2001comparing}.
The KSC set is created by mixing samples from two different source corpora $A$ and $B$ in gradually-changing proportions. 
The KSC set, denoted $KSC(A,B)$, consists of $k$ corpora $\{c_1,c_2,\dots,c_k\}$, each of size $n\geq k-1$, where corpus $c_i,\:i=1,\dots,k$ is constructed by sampling $n\left(\frac{k-i}{k-1}\right)$ observations from $A$, and the remaining $n\left(\frac{i-1}{k-1}\right)$ from $B$ (see Figure \ref{fig:ksc}). 
The sampling resolution gradation between corpora is a fixed $\frac{1}{k-1}$.


We now introduce some notation on the $KSC$ set, which are used to define the measures in Section~\ref{sec:approach}.
Let $[k]=\{1,2,\dots,k\}$.
For given source corpora $A$ and $B$, for each $\ell\in[k-1]$ we define the \textbf{$\ell$-distant corpora set} as follows:
\begin{equation}
    KSC_\ell(A,B) = \left\{(c_i,c_j)\colon\:i,j \in [k],\: j-i=\ell\right\}
\end{equation}
Let $d(a,b)$ denote the distance from corpus $a$ to $b$, according to metric $d$.
Let $D_\ell (A,B,d)$--- $D_\ell$ for short---be the set of values of distance $d$ for corpora pairs in $KSC_\ell(A,B)$;  
\begin{equation}
\vspace{-0.5em}
    D_\ell(A,B,d)=\{ d(c_i,c_j)\colon\: (c_i,c_j) \in KSC_\ell(A,B)\}    
\end{equation}
To pool results across $\ell$, we further define:
\begin{equation}
D(A,B,d)= \{D_\ell (A,B,d)\colon\: \ell \in [k-1]\}
\end{equation}
Note that because we do not require the distance metrics considered to be `metrics' in the mathematical sense, they may not be symmetric (i.e., possibly $d(a,b)\ne d(b,a)$).  However, since $KSC_{\ell}$ enforces a pairwise order on pairs $(c_i,c_j)$ by requiring $j>i$, this ensures that $D_\ell(A,B,d)$ is properly defined.

Some of the metrics $d$ have a pre-defined range (e.g., CHI, MAUVE, DC, PR only return values in the range [0,1]) while others have no preset scale or operation range.
Therefore, to allow sensible comparison of distance metrics with different operation ranges and across source corpora, we obtain $z$-scores by normalizing the metric values, pooled across all $D_\ell(A,B,d)$. 
In the following analysis, if not specified otherwise, $D_\ell$ will always be the normalized rather than raw distances.

\paragraph{Datasets Selection}
The measures described in Section~\ref{sec:approach} are applicable to any pair of textual datasets with differently-distributed textual content, allowing the corpora in the KSC set to be distinguishable.
To ensure that each pair of source corpora were in fact different enough, in the following experiments we use pairs of human text corpora from different domains, rather than pairing a human corpus with a machine-generated version of itself.
For our experiments we selected four public datasets (ATIS, \citealt{atis}; yahoo\footnote{https://ciir.cs.umass.edu/downloads/nfL6/}; banking77, \citealt{banking77}; clinc150, \citealt{larson2019evaluation}) containing short user utterances from different domains summarized in Table \ref{tab:datasets}.

\begin{table}
\centering
\resizebox{1\columnwidth}{!}{
\begin{tabular}{lll}
\hline
\textbf{Name} & \textbf{Size} & \textbf{Description}\\
\hline
atis & 4978 &  Utterances to a flight  \\
& & booking system. \\
yahoo & 20000 & Yahoo non-factoid\\
& & questions in 21 categories. \\
clinc150 & 22500 &  Utterances in 10 domains\\
& &  classified into 150 classes. \\
banking77 & 10000 & Online banking queries.\\
\hline
\end{tabular}}
\caption{
Datasets used as source corpora in our benchmark. Although some of the datasets are partitioned annotated with labels, in our experiments, if not mentioned otherwise, we ignored those labels.
}
\label{tab:datasets}
\end{table}


\section{Metric Robustness Measures}
\label{sec:approach}



\begin{figure*}
    \centering
    \includegraphics[width=1\textwidth]{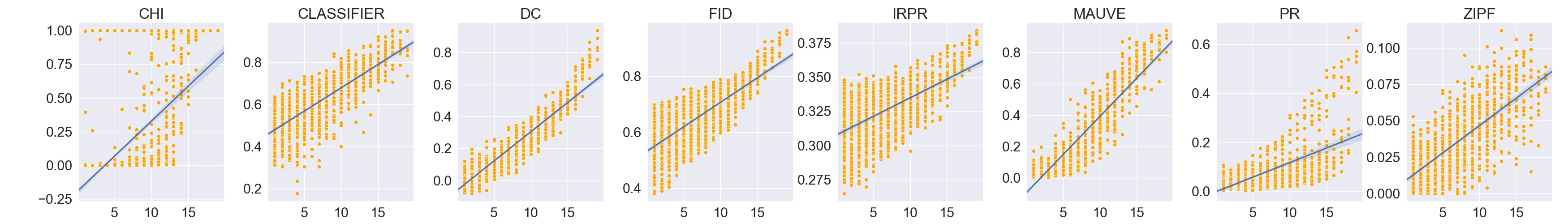}
    \includegraphics[width=1\textwidth]{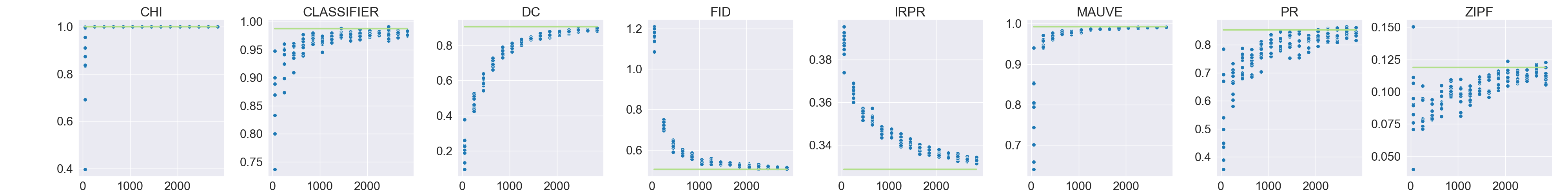}
    \includegraphics[width=1\textwidth]{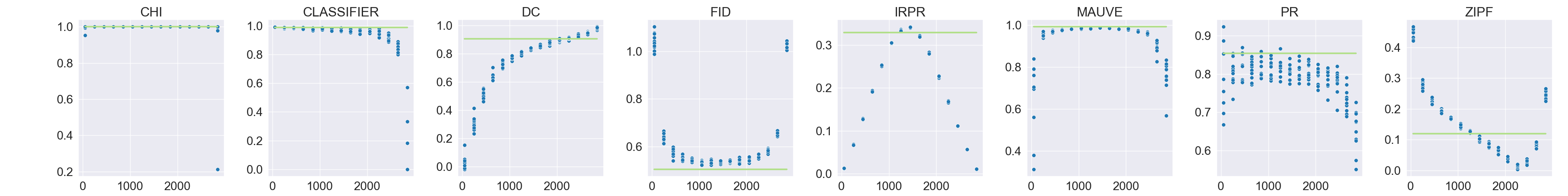}
    \vspace{-1.5em}
    \caption{\label{fig:scatters}
    Top: Distance values (non-normalized) of corpora pairs in $D_{\ell}$ versus $\ell$. ($n=100, k=12, |J|=6053$), pooled across 5 repetitions of KSC samples.
    Blue line indicates regression and confidence interval at 95\%. 
    Middle: Distance values calculated on increasing $s$ size corpora $a_s$ and $b_s$ sampled from sources $A$ and $B$,  correspondingly. 
    Bottom: Distance between imbalanced corpora $a_s$ and $b_{\bar{s}}$ where $|b_{\bar{s}}|=N-s$ and $N=2900$. 
    The x-axis represents $s\in\{50,250,450,\dots,2850\}$  ($repetitions=10$).
    In middle and bottom figures, green horizontal line indicates the asymptotic distance $d(A,B)$.
    In all figures $A$=clinc150 and $B$=banking77.
    }
\end{figure*}

We now describe our measurements of desirable properties for distance metrics, given the normalized $D_\ell$ on the KSC sets. 
In the three following measures (Monotonicity, Separability, and Linearity), we aim to capture three attributes of well-behaved metrics that can be understood by considering the top line scatter plots of Figure \ref{fig:scatters}; these show the relation between the $D_\ell$ sets and $\ell$.
In these scatterplots, a high angle of the regression line, low vertical variability  around it, and linearity are all desirable properties for the distance metric, and are captured in these measures.


\subsection{Metric Monotonicity}
A well-behaved distance metric $d$ should have a natural monotonic relationship with the separation levels $\ell$ of the KSC.
We use Spearman's rank correlation between $\ell$ and $D_\ell$, which we denote $\rho(d)$, to assess the monotonicity.
Spearman's correlation is defined as the Pearson correlation between the order ranks of two variables, and measures the strength of their monotonic, rather than linear, association.
As can be seen in Table \ref{tab:results}, MAUVE and CHI achieve the best monotonicity results, followed by DC and FID.



\subsection{Metric Separability}
It is desirable that (1) for a given $\ell$, $D_\ell$ has low variability, and (2) for different $\ell_2>\ell_1$, the samples $D_{\ell_1}$ and $D_{\ell_2}$ are distinguishable (e.g., by a two-sample difference test), particularly as $\ell_2-\ell_1$ grows. 
Here, we measure how grouping by $\ell$ explains the variability in $D_{\ell}$ across $\ell$.
We perform a one-way fixed-effects analysis of variance (ANOVA) with $\ell$ as the unordered categorical treatment and $D_\ell$ as the numeric response. 
Often, an F-test is performed; if its p-value is low, it means a significant amount of the variance in the response ($D_\ell$) can be explained by the treatment ($\ell$). 
Since the F-test for any reasonable $d$ metric should be significant, we instead use the similar $\omega^2$ effect-size metric \citep{hays1963statistics}, which is bounded by $\pm 1$, to better assess them. 
It is defined as
\begin{equation}
    \omega^2=\frac{SS_{\textrm{treatment}}-df_{\textrm{treatment}}\times MS_{\textrm{residual}}}{SS_{\textrm{total}}+MS_{\textrm{residual}}}
\end{equation}
where $SS$ and $MS$ are the sum and mean sums of squares, and $df$ is the degrees of freedom, on a dataset of size $n$ (here, $n=|D(A,B,d)|$).  
In the following we denote this measure as $\mathcal{W}(d)$.

\subsection{Metric Linearity}
Here we examine to what extent linear changes in the corpus content ($\ell$) are manifested in linear changes in the distance function.
To do so, we  calculate the coefficient of determination ($R^2$), where higher values indicate stronger linearity.
This measure is denoted by $\mathcal{L}(d)$.
Looking at the results in Table \ref{tab:results}, we see that MAUVE achieves the best results followed by DC and FID.
It appears that this measure is more affected by the source corpora and by the resolution than other metrics.




\subsection{Metric Time Efficiency}
The time complexity of the metric is commonly perceived as less important, thus seldom reported \cite{sai2022survey}. 
This aspect is becoming ever more important, especially due to the growing interest in time-consuming divergence frontier methods \cite{djolonga2020precision}.
Such metrics perform multiple measurements to estimate the area under the curve (similar to precision-recall curves for binary classification), with tune-able but increasing resolution.
We measure the time performance of the metric $\mathcal{T}(d)$ in terms of $100$ similarity measurements operations per second ($[100\emph{op}/sec]$) on a standard CPU machine\footnote{CPU: 2.3 GHz 8-Core Intel Core i9. Memory: 64 GB DDR4 (2667 MHz)}.
Note that the measurements reported in Table \ref{tab:results} do not include the sentences' embedding time.
Predictably, methods that operate on the token level and avoid complex density estimation tend to achieve the best time performance.
Among the distributional metrics, MAUVE achieves the best results, followed by FID.
PR and DC produce similar results since both are based on similar manifold calculations.

\begin{table*}
\centering
\resizebox{1\linewidth}{!}{
\begin{tabular}{lr|lr|lr|lr|lr|lr|l||cc}
\hline
 & \multicolumn{2}{c}{$\mathcal{A}(d)$} & \multicolumn{2}{c}{$\mathcal{A}^w (d)$} & \multicolumn{2}{c}{$\mathcal{T}(d) $} & \multicolumn{2}{c}{$\rho(d)$} &  \multicolumn{2}{c}{$\mathcal{W}(d)$} & \multicolumn{2}{c}{$\mathcal{L}(d)$} & $\mathcal{S}(d)$ & $\mathcal{I}(d)$\\
 $k$& $7$ & $12$ & $7$ & $12$ & $7$ & $12$ & $7$ & $12$ & $7$ & $12$ & $7$ & $12$ & & \\
\hline

CHI        & .945          & .852          & .913          & .774          & \textbf{4.68} & \textbf{3.29} & .875          & .866          & .684          & .702          & .810           & .767          & \textbf{.989} & \textbf{.994} \\
CLS. & .789          & .701          & .731          & .618          & 1.26          & 1.10          & .704          & .735          & .544          & .562          & .767          & .767          & .972          & .918          \\
DC         & .958          & .863          & .936          & .805          & .031          & .031          & .913          & \textbf{.892} & \textbf{.908} & \textbf{.879} & \textbf{.946} & \textbf{.919} & .832          & .808          \\
FID        & .949          & .810           & .923          & .753          & .067          & .066          & .764          & .695          & .563          & .537          & .81           & .759          & .821          & .877           \\
IRPR       & .832          & .710           & .784          & .638          & 4.39          & 2.35          & .571          & .543          & .258          & .275          & .645          & .598          & .949          & .856          \\
MUV.      & \textbf{.976} & \textbf{.888} & \textbf{.963} & \textbf{.828} & .079          & .071          & \textbf{.938} & \textbf{.906} & .883          & \textbf{.885} & \textbf{.947} & \textbf{.926} & \textbf{.977} & .943          \\
PR         & .820           & .688          & .767          & .608          & .031          & .031          & .649          & .592          & .577          & .566          & .716          & .667          & .909          & .934            \\
ZIPF       & .886          & .726          & .851          & .657          & 4.65          & 2.668          & .751          & .633          & .514          & .413          & .785          & .667          & .852            & .913        \\

\hline
\hline

CHI        & .953          & .935          & .936          & .891          & \textbf{5.58} & \textbf{3.33} & \textbf{.960} & \textbf{.962} & .835          & \textbf{.900}   & .83           & .858         & \textbf{1.00}               & \textbf{1.00}               \\
CLS. & .931          & .827          & .902          & .751          & 1.29         & 1.21          & .843         & .836          & .671          & .702          & .847          & .847         & \textbf{.993}  & \textbf{.989} \\
DC         & .773          & .601          & .702          & .552          & .031          & .031           & .707         & .615          & .763          & .717          & .825          & .759         & \textbf{.988}  & \textbf{.986} \\
FID        & .967          & .904          & .947          & .848          & .071          & .067           & .793         & .754          & .634          & .636          & .845          & .816         & .922           & .898          \\
IRPR       & .697          & .587          & .642          & .570           & 3.91         & 2.69          & .382         & .264          & -.02          & -.001         & .433          & .341         & .951          & .834          \\
MUV.      & \textbf{.999} & \textbf{.977} & \textbf{.998} & \textbf{.961} & .084          & .067           & .936         & .943          & \textbf{.856} & \textbf{.904} & \textbf{.932} & \textbf{.950} & \textbf{.999} & \textbf{.994} \\
PR         & .722          & .467          & .666          & .446          & .031          & .030            & .459         & .240           & .488          & .394          & .658          & .523         & .890          & .899          \\
ZIPF       & .854          & .783          & .817          & .736          & 4.77         & 3.02           & .660          & .635          & .309          & .352          & .687          & .661         & .735          & .904    \\
\hline
\end{tabular}}
\caption{
Summary of metrics evaluation scoring on two pairs of source datasets in low ($k=7$) and high ($k=12$) resolution KSC ($n=100$). 
Best results with differences below $.015$ are marked in bold.
$\mathcal{T}(d)$ units are $[100 \emph{op}/sec]$.
MUV. stands for MAUVE and CLS. for CLASSIFIER. 
In the top table, $A$=clinc150 and $B$=banking77. In the bottom table $A$=atis and $B$=yahoo. 
The average results of $5$ repetitions are reported for all measures except size and imbalance robustness, in which the number of repetitions is $10$. 
More statistical details are provided in Figure \ref{fig:ksc_additional_reasults} in the Appendix.}
\vspace{-1.5em}
\label{tab:results}
\end{table*}

\subsection{Metric Accuracy \label{ssec:metric_accuracy}}

The assessment measures described earlier in Section~\ref{sec:approach} use the observed values of the metric distances (or similarities) between the KSC corpora; however, the actual values of the distance may not be known.
Nevertheless, we still have some partial information about the ordering of these values, which we will use to define an accuracy measure, which requires us to define the notion of a `judgement', as follows.

\subsubsection{Comparing paired corpora distances \label{sssec:judgements}}

Suppose we do not know the observed \textit{values} of $d$ in $D(A,B,d)$ for the paired corpora in $KSC_{\ell}(A,B)$, pooled across $\ell$.
Nevertheless, we can still assume that certain pairwise distances are larger than others.
For instance, the proportions of observations from $A$ in $c_2$ and $c_3$ are more similar than the respective proportions between $c_1$ and $c_4$.  
Moreover, the interval of the first pair is `contained'\footnote{$(c_i,c_j)$ contains $(c_q,c_r)$, i.e., $(q,r)\subset(i,j)$, if $i\leq q$ and $r\leq j$ and $i<r$.} in the second, and thus the first pair should have smaller distance. 
Thus, it should be true that, say, $d(c_2,c_3) \leq (c_1,c_4)$ in expectation (across repeated random sampling).
In general, whenever the interval of one corpus pair contains ($\subset$) the interval of another, we expect the contained pair to have a smaller distance.

\begin{figure}[h]
    \centering
    \includegraphics[width=0.4\textwidth]{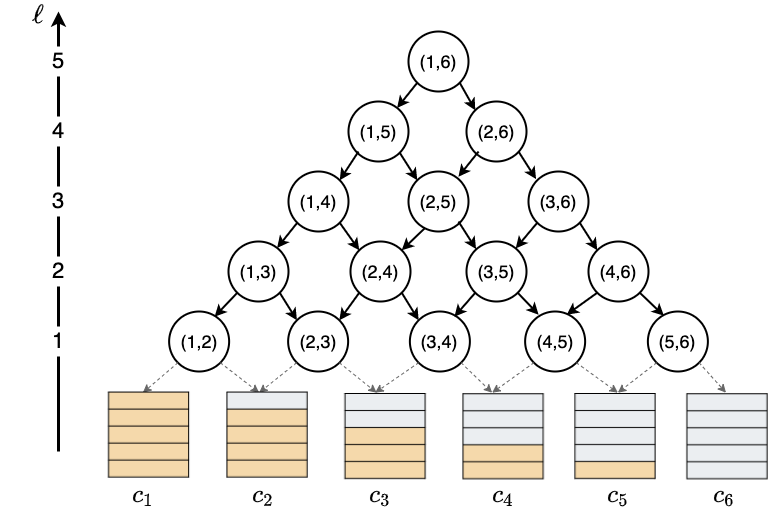}
    \caption{A tree representation of the judgements performed on the KSC collection given a metric $d(\cdot,\cdot)$, for calculating the accuracy ($\mathcal{A}$, Section~\ref{ssec:metric_accuracy}) measures. 
    The leafs are the KSC collection and the inner nodes (circles) represent the corpora tuples ($c_i,c_j)$.  
    The set $J$ contains all judgements such that each node $(i,j)$ is judged against all descended nodes.
    Namely, if there is a path from node $a$ to node $b$, there is a judgement between the two nodes, and the judgement is correct if $d(b)\leq d(a)$. The size of the judgements set can be expressed as:
$|J|=\sum_{i=1}^k (k-i){\left( \frac{i(i-1)}{2} -1 \right)}$.  For instance, $|J|=339$ if $k=7$, and 6053 if $k=12$.}
    \label{fig:ksc_tree}
\end{figure}

Given two pairs, $(c_i,c_j)$ and $(c_q,c_r)$, of paired corpora, we can only reliably predict\footnote{For instance, say we compare pairwise distances between $(c_1, c_6)$ and $(c_5, c_7)$.
Even though the second interval length ($7-5=2$) is smaller than the first ($6-1=5$), because it is not contained in the first, we cannot necessarily say that $d(c_5,c_7)\leq d(c_1,c_6)$ since inter-corpus distance may not be proportional to the interval length.} which of $d(c_i,c_j)$ or $d(c_q,c_r)$ is larger in expectation (a decision we call a `judgement') if the interval of one pair contains the other's.
The set $J$ contains all and only such judgements:
\begin{equation}
\begin{split}
J=\{((c_q,c_r),(c_i,c_j)) \colon (q,r)\subset(i,j)\}
\end{split}
\end{equation}
The judgement $d(c_q,c_r)\leq d(c_i,c_j)$ is correct when the second interval contains the first.
This gives the most probabilistically-logical partial order on the similarities between corpora in a KSC collection, that can be obtained without knowledge of the true pairwise $d$-distances between corpora.
Figure \ref{fig:ksc_tree} shows a tree representation of KSC-set pair containment relations, from which the set of judgements $J$ can be extracted.

\subsubsection{Accuracy \label{sssec:accuracy}}

The metric accuracy is defined as the rate of correct judgements, formally:
\begin{equation}
\mathcal{A}(d)=\frac{1}{|J|}\sum_{\jmath \in J} \mathds{1}( d(c_q,c_r)\leq d(c_i,c_j))
\end{equation}
where $\jmath=((c_q,c_r),(c_i,c_j))$ is a judgement in $J$ and $\mathds{1}(\cdot)$ is the indicator function.
Further, we propose a weighted version of the accuracy metric that assigns more weight to harder judgements.
We define the hardness of judgement $\jmath$ as $w(\jmath)=\frac{1}{\ell_2-\ell_1}$ where $\ell_2=j-i$ and $\ell_1=r-q$, and $\ell_2>\ell_1$ by definition of $J$.
Formally,
\begin{equation}
\mathcal{A}^w(d)=C \sum_{\jmath \in J} w(\jmath) \cdot \mathds{1}(d(c_q,c_r) \leq d(c_i,c_j)))
\end{equation}
where $C=\left(|J| \cdot \sum_{\jmath \in J}w(\jmath)\right)^{-1}$.
While $\mathcal{A}$ and $\mathcal{A}^w$ are correlated, as one may expect, $\mathcal{A}^w$ typically returns lower values (see Table \ref{tab:results}).

In our implementation, the set of samples in each $c_i$ is disjoint,  namely, $c_i \cap c_j =\emptyset, \forall c_i,c_j \in KSC(A,B)$.
This was done to prevent perfect judgements by naively counting the number of common instances (e.g., by defining $d(c_i,c_j)=|c_i \Delta c_j|$ where $\Delta$ denotes the symmetric difference).
MAUVE, followed closely by FID, CHI and DC, achieves the highest accuracy results across resolutions and source corpora.

\subsection{Size Robustness}
\label{subsec:size_robustness}
We are also interested in capturing the sensitivity of a metric to sample sizes. 
To accomplish this, we need to quantify the convergence pace of $d(a_s,b_s)$ to the asymptotic distance $d(A,B)$, where $a_s,b_s$ are samples from corpora $A,B$ of increasing size $s$. 
Specifically, in our experiments $s \in S=\{50,250,450, \dots, 2850\}$.
The middle plot in Figure \ref{fig:scatters} shows convergence patterns of the different metrics to the asymptotic distance.
The asymptotic distance is estimated by the mean of repeated ($10$) calculations of the distance on samples of size $3000$ each from $A,B$, rather than on the full corpora.
To quantify the metric size robustness, $\mathcal(S)$, we calculate the mean absolute error, $|d(a_s, b_s)-d(A,B)|$, for all $s\in S$, normalized by the asymptotic distance:
\begin{equation}
    \mathcal{S}(d) = 1-\sum_{s \in S} \frac{|d(a_s, b_s)-d(A,B)|}{d(A,B)}
\end{equation}
Similar to previous measures, the normalization is performed to omit the influence of metric scale and operation ranges.

While our results demonstrate (Figure \ref{fig:scatters}) that most of the metrics examined require around $1000$ samples to closely estimate the asymptotic distance between the source corpora, their measured accuracy ($\mathcal{A}$(d) and $\mathcal{A}^w(d)$) is still fairly high even on small corpora within the $KSC$, and can capture relative differences in corpus content.


\subsection{Imbalance Robustness}
Similarity metrics are frequently used to compare datasets with unequal sample size. 
Especially when comparing real and generated corpora, the size of a generated corpus is usually much larger than the real corpus.
The imbalance robustness measure quantifies the effect of corpora size imbalance on the metric's performance (see Figure \ref{fig:scatters}, bottom).

Unsurprisingly, asymmetric metrics such as PR and DC are most affected by size imbalance.
While PR, DC, and MAUVE were all originally designed to measure the disparity between human and generated data (and thus asymmetric in the reference$P$ and target $Q$), it seems that MAUVE overcomes the sensitivity to datasets of very unequal sizes.
Interestingly, imbalance causs some metrics (CLASSIFIER and MAUVE) to underestimate the distance, while others  (FID) overestimate it.
When we compare the convergence patterns of PR and DC, both are similarly asymmetric, maintaining $d(P,Q)$. When we increase the reference size, PR diverges from the true asymptotic distance, while DC converges to it.
The Imbalance Robustness score $\mathcal{I}(d)$ is calculated similarly to the size robustness score, only that $|b_s|=N-|a_s|$.

\begin{figure}[ht!]
    \centering
    \includegraphics[width=0.95\columnwidth]{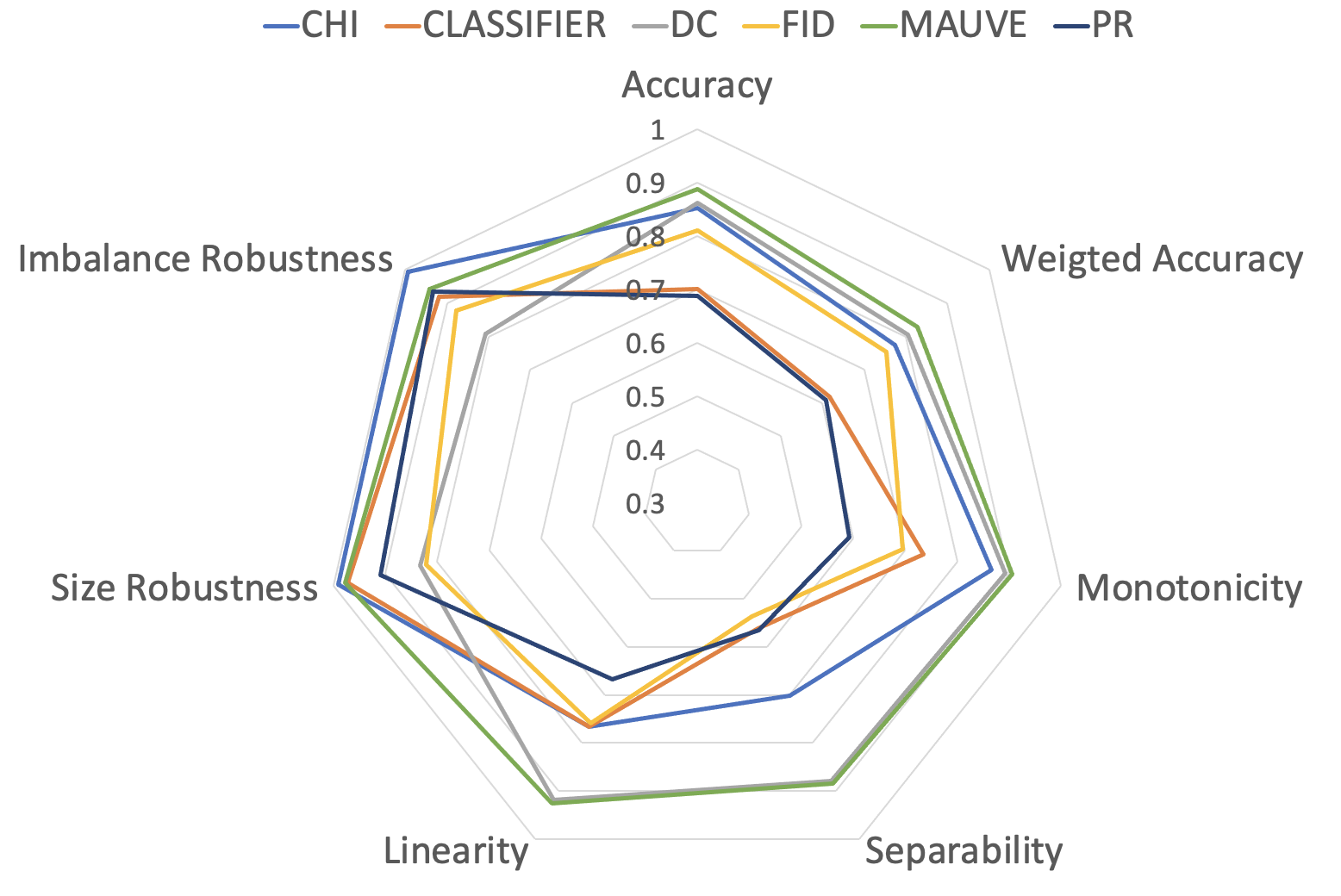}
    \caption{Leading metrics characterization radar chart. Mean results from Table \ref{tab:results} for $A$=clinc150,  $B$=booking77 and for $k=12$, excluding time efficiency to maintain scale.
    } 
    \label{fig:radar_atis_yahoo_high}
\end{figure}



\begin{figure*}[h!]
    \centering
    \includegraphics[width=1\textwidth]{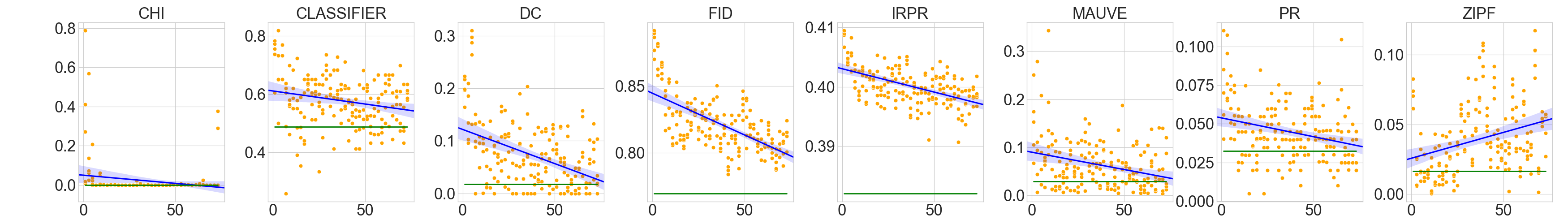}
    \includegraphics[width=1\textwidth]{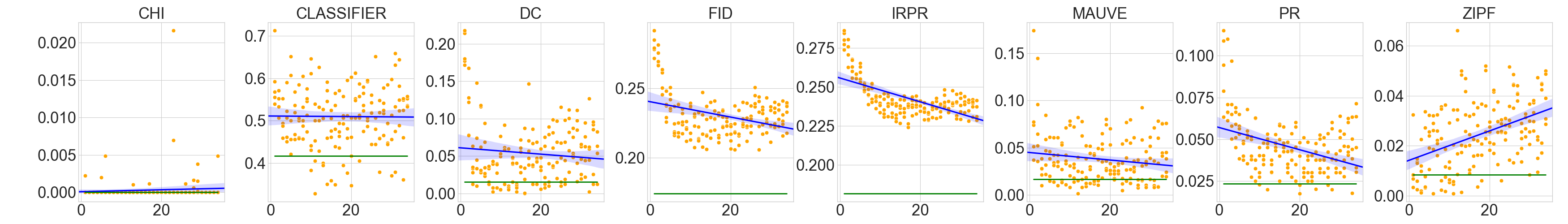}
    \caption{
    Similarity between reference corpus and iteratively fine-tuned corpora $g_i$ samples.
    Orange dots show the similarity between samples of generated text in iteration $i$ and the source dataset. 
    The blue line indicates regression and confidence interval at 95\%.
    The green horizontal line specifies the mean estimation of the distance between two random samples of the original corpus. 
    The top figure shows iterative generation on unlabeled news headlines dataset.
    The bottom shows the iterative conditional generation using LAMBADA \cite{anaby2020not} trained on banking77 dataset.
    }
    \label{fig:generation_trends}
    \vspace{-1em}
\end{figure*}

\paragraph{KSC Parameters}
As shown in Section \ref{subsec:size_robustness}, most metrics require at least $n=1000$ samples to capture the true distance between two source domain corpora; however, our experiments use $n=100$.
This is because our measures are relative, i.e., we do not aim to calculate the true asymptotic distance between two domains, but to measure the metrics' robustness in detecting small changes in the compared corpora.
Furthermore, if $n$ is large, $k$ must also be large to ensure the $k$ corpora in the KSC set have small enough absolute consecutive differences.

\noindent
Note that small consecutive differences in KSC corpora are needed so that the measures in Section~\ref{sec:approach} will have a high enough resolution and large enough sample size of $D_{\ell}$ to properly differentiate the metric properties.
In particular, this ensures the judgements (Section~\ref{sssec:judgements}) used in the accuracy measures ($\mathcal{A}$ and $\mathcal{A}^w$) are not too `easy' to make correctly, in which case they would be less useful as a tool.
For instance, a metric with 100\% accuracy makes all correct judgements, e.g., that $d(c_2,c_3)\leq d(c_1,c_4)$.
If $k=5$, the gap (in expectation) between the pair distances compared is large, so the judgement is easy, and  thus all metrics may have full accuracy.
When $k$ increases, the absolute consecutive differences in corpora fall, and thus the difficulty of the judgement increases.
Some metrics will fail to make the  judgement correctly (in a given random KSC), decreasing their accuracy; this allows us to better differentiate between the more and less accurate metrics.
However, setting $k$ too high results in a computationally prohibitive number $|J|$ of judgements.
Therefore, we opted to use the smaller $n$ that are still sufficient to capture the quality and robustness of the investigated metrics.\\

\section{Increasingly Fine-tuned Corpora}
Here, we qualitatively investigate the metrics' ability to discriminate between generated and human text using the following procedure:
We generated a sequence of equal-size synthetic corpora $IFC = (g_1, g_2,\dots, g_n)$ by sampling from a gradually fine-tuned language model on a specific source corpus $A$.
Namely, in each iteration, a fine-tuning step is performed by training the language model on a single epoch containing $1000$ sentences randomly drawn from $A$, followed by a generation process to synthesize a corpus $g_i$ containing around $1000$ sentences.
The name IFC, or "Increasingly Fine-tuned Corpora", was chosen to parallel the name KSC ("Known Similarity Corpora").

For each generated corpus $g_i$, we estimated the distance from $A$, i..e, $d_i=d(A,g_i),\: \forall i\in [n]$.
While the true distance between those synthetic corpora and $A$ are unknown, an effective metric should capture the decreasing distance between $A$ and $g_i$ with increasing $i$, namely $d_1\prec d_2\prec\dots \prec d_n$.
Due to our results, which show low imbalance robustness of some metrics, we maintained the same-size corpora when calculating corpora distance.



The results presented in Figure \ref{fig:generation_trends} show the gap between human and generated text captured by each metric in each iteration.
To calculate the average self-distance of the reference corpus ($A$), we take the mean distance between two randomly sampled sub-corpora $r_1$ and $r_2$ from $A$, i.e. $d(A,A)=\mathds{E}
_{r_1,r_2 \sim A}[d(r_1, r_2)])$.

In our experiments we used two datasets, the banking77 dataset, mentioned above and the news dataset\footnote{HuffPost (www.huffpost.com) news headlines collected from 2012 to 2018 containing around 200k headlines.(www.kaggle.com/rmisra/news-category-dataset (https://www.huffpost.com)}, representing different domains of text corpora.
The IFC set for the banking77 dataset was generated in an iterative two-step procedure similar to the one described in LAMBADA \cite{anaby2020not}. 
This procedure first generates sentences conditioned on the label, then filters out sentences that are out-of-domain or incorrectly labeled. 
However, the IFC set for the news dataset was generated by finetuning the pre-trained GPT-2 medium  model \cite{radford2019language}.

The results in Figure \ref{fig:generation_trends} show that CHI is less effective than the other metrics in capturing the gradual nature of the $IFC$.
Also, they show that FID and IRPR are sensitive in discriminating between the original and generated corpora, even after many fine-tuning iterations.
Interestingly, the ZIPF distance increases with the iteration.
This indicates that the generated text, despite becoming semantically closer to the original with the increasing  iterations, becomes less `natural' in that the token frequencies deviate from that of human text and the reference corpus.
This can be explained, at least in part, by the TTR measure. TTR is a standard word diversity measure, calculated by dividing the number of unique words in a text by the total word count. A high TTR indicates significant lexical variation.
Indeed, in the IFC of banking77, $g_1$'s TTR is 0.295 which is closer to the original dataset's TTR of 0.299 than $g_{40}$'s TTR of 0.322.

\section{Conclusions}
\label{sec:conclusions}
In this work, we propose a principled set of automatic measures for evaluating the quality of text dissimilarity metrics.
By testing various metrics using our measures, we show that they do a good job of capturing their known characteristics, hence increasing our confidence in these measures; also, overall, recent metrics exhibit more favorable traits than their predecessors. 
The radar chart in Figure \ref{fig:radar_atis_yahoo_high} shows that our measure scores correlate well with the compared distributional metrics recency $MAUVE \succ DC \succ PR \sim FID$ as well as their known relative strengths.


\section{Limitations}

Although one of the main motivations for comparing corpora is to measure the semantic gap between human and generated short text, we used pairs of human text corpora from different domains to maintain controllably-distinct corpora in the KSC set.
Despite this, future efforts to develop human and machine-generated benchmark pairs \cite{mille2021automatic} will allow for future work to quantitatively measure the characteristics of semantic metrics on pairs of human and generated corpora using the approach devised in this paper.


Also, for more straightforward comparisons, we used only a single sentence-embedding model.
However, as other studies (e.g., GPT-2 \cite{radford2019language} in \citet{pillutla2021mauve} and Bert \cite{devlin2018bert} in \citet{lo2019yisi}) have shown, the quality of a corpus distance metric can be affected by the embedding choice.
In future extensions of our work, we plan to allow for multiple embeddings to obtain a more refined evaluation of the metrics. 

An important limitation of this work is that it considers only English corpora of short text samples. 
We examined only a limited set of metrics and datasets, both of which we intend to extend. 

In addition, we note that while our experiments calculate all KSC-based measures using a single KSC collection (same $n$ and $k$ values), it could be favourable to use different $n$ and $k$ for different measures. 
For instance, the time performance is calculated using a single size small dataset $n=100$. 
In future work, the time scalability of metrics can be more closely investigated by comparing their time performance on increasing corpora sizes.

As indicated in Section \ref{sec:approach}, creating KSC collections with large $k$ creates an excessive number of judgements (e.g., for $k>15$, $|J|>50000$), thus limiting the scalability of our method to smaller $k$ and thus smaller $n$, if high resolution is required.
This would preclude comparing the robustness of metrics that require large samples.
We intend to rectify this in future work by creating representative smaller judgement sets by carefully sampling from the complete set. 

As mentioned in Section \ref{sec:literature}, some of the investigated metrics were adapted to return a single value summarizing the distance between two corpora (e.g., averaging the precision and recall by the $F1$ score).
Further work is required to build measures that can compare metrics returning multiple values.

\bibliographystyle{acl_natbib}
\bibliography{main}

\appendix
\include{AppendixA}

\end{document}

%% file: AppendixA.tex
\onecolumn
\section{Appendix}
\label{sec:appendix}

\begin{figure*}[h!]
    \centering
    \includegraphics[width=1\textwidth]{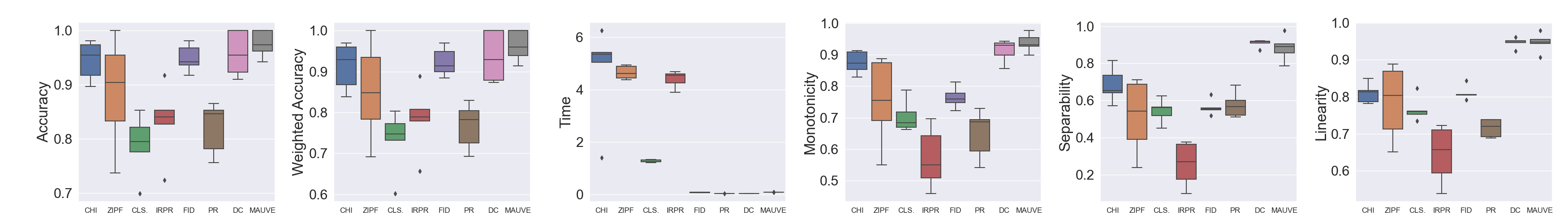}
    \includegraphics[width=1\textwidth]{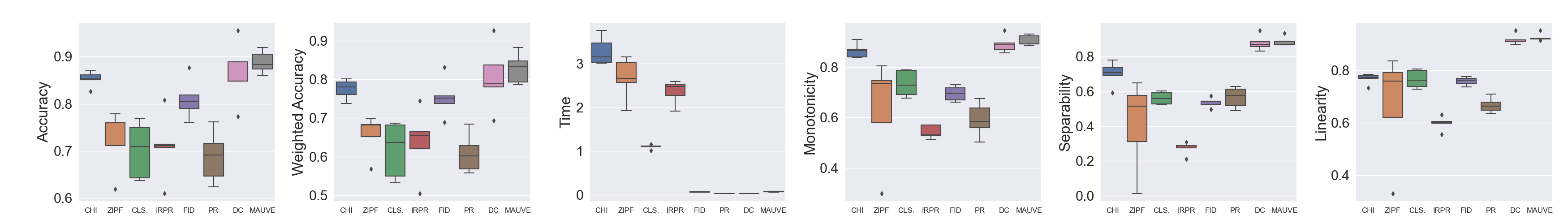}
    \includegraphics[width=1\textwidth]{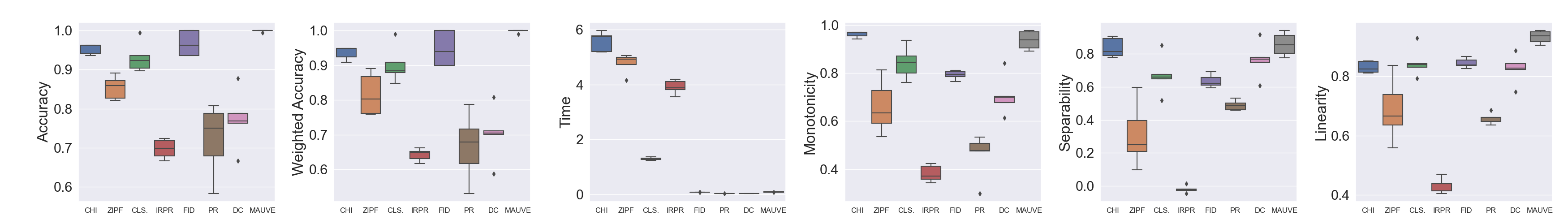}
    \includegraphics[width=1\textwidth]{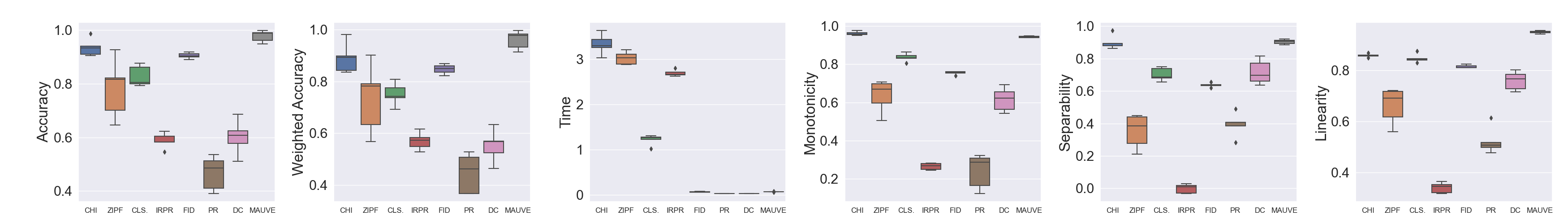}
    \caption{Distributional information of the results shown in Table \ref{tab:results}. The top two figures showing the results for ($A$=clinc150 $B$=banking77), for $k=7$ and $k=12$, respectively.
    The bottom two figures are for ($A$=yahoo $B$=atis), for $k=7$ and $k=12$, respectively.
    Colored boxes depict the interquartile ($25^{th}$ to $75^{th}$) range. 
    The mean is indicated by a horizontal line.
    All data points within 1.5 of the corresponding limits of the interquartile range are depicted by whiskers. Data points outside this range are plotted individually.
    CLS. indicates the CLASSIFIER metric.}
    \label{fig:ksc_additional_reasults}
\end{figure*}